\documentclass[11pt,letterpaper]{article}

\usepackage[margin=1in]{geometry}

\usepackage{graphicx} 
\usepackage{booktabs}
\usepackage{tabularx}
\usepackage[table]{xcolor}
\usepackage{array}
\usepackage{amsmath}
\usepackage{float}
\usepackage{url}

\usepackage{censor}

\title{\LARGE \bf
Mediating Human-Swarm Interactions Through a Robot Helper:\\ A Shepherding-Inspired Approach
}

\author{Arihant Yadav$^{1}$, Sophie J\"{o}rg$^{2}$ and Eakta Jain$^{1}$
\thanks{$^{1}$Arihant Yadav \& Eakta Jain are with Department of Computer \& Information Science \& Engineering
,
        University of Florida, Gainesville, USA
        {\tt\small arihant.yadav@ufl.edu, ejain@ufl.edu}}
\thanks{$^{2}$Sophie J\"{o}rg is a Professor in the Information Systems and Applied Computer Sciences Department at University of Bamberg,
       Germany
        {\tt\small sophie.joerg@uni-bamberg.de}}%
}

\begin{document}

\maketitle
\thispagestyle{empty}
\pagestyle{empty}

\begin{abstract}
This work investigates human-swarm interaction (HSI) for steering large collectives through an intermediary robot helper. While prior robot-swarm control and HSI techniques commonly assume swarm members are addressable, these assumptions often fail for ``ungovernable'' swarms such as wildlife flocks and schools of fish. Motivated by pastoral shepherding, we compare low-level teleoperation to a higher-autonomy motion-primitives interface in a virtual reality (VR) shepherding task under varied swarm cohesion dynamics. Results indicate that motion primitives that operate on collective shape (e.g., boundary-based control) improve task performance and reduce perceived workload. 
\end{abstract}

\section{INTRODUCTION}
Large, naturally occurring collectives such as bird flocks and fish schools often exhibit the properties of \textit{ungovernable swarms}: systems in which individual members cannot be pre-programmed or directly commanded, yet whose aggregate motion must still be influenced for safety or logistical reasons (e.g., airfield operations, aquaculture).
At present, influencing of such swarms is largely achieved through passive or environmental interventions. In aviation, for example, wildlife hazard management commonly prioritizes habitat modification (e.g., reducing attractants such as food, water, and shelter) and exclusion methods to discourage birds from using the airfield environment~\cite{ClearyDolbeer2005,Blackwell2009LandUseAirports}. Only under particular contingencies do airports more frequently deploy reactive dispersal and direct-control tactics (e.g., hazing/harassment, pyrotechnics, or removal)~\cite{BelantMartin2011,ACRP32_2010}. Although effective in many routine contexts, these approaches are difficult to apply with precision during time-critical disruptions.
A canonical example where an ``ungovernable'' collective is influenced via a human-controlled helper agent is pastoral shepherding. Here, a handler issues compact, discrete commands (e.g., ``come-bye'', ``walk-up'') to a sheepdog. The sheepdog then transmits the human’s intent by shaping flock-level motion. This paradigm has inspired a substantial body of algorithmic and robotic shepherding research, in which one or more agents guide swarm members~\cite{Strombom2014,Long2019}.

Translating shepherding into human-robot interaction (HRI) raises a central interface-design question: \textbf{how should the human’s interaction with the helper agent be structured?} The designer must decide how much of the helper’s behavior should be specified explicitly by the human versus generated autonomously by the robot, and at what level of abstraction (e.g., continuous low-level control versus discrete, task-level commands). This choice directly determines the robot’s effective autonomy and, consequently, the operator’s role in the control loop.
In HRI, ``levels of robot autonomy'' (LoRA) are consequential because they shape operator workload, situation awareness, and performance. Beer et al. formalize this idea via a 10-level taxonomy spanning teleoperation through full autonomy, enabling systematic reasoning about autonomy allocation and its human-factors implications~\cite{Beer2014}. 

In our setting, this autonomy-allocation question manifests concretely as an interface design choice: whether the human should continuously control the helper agent at the motion level, or instead delegate execution details to the agent through higher-level commands. Guided by the shepherding metaphor, we compare (i) a teleoperation condition, which provides direct, continuous control of the helper agent, with (ii) a shepherding condition with a motion-primitives based interface. This comparison is motivated by prior work on shared autonomy and primitive-based control which argues that elevating control from low-level actuation to higher-level task/motion primitives can reduce operator burden and improve efficiency while preserving meaningful human agency~\cite{Javdani2018,Gopinath2016,ijspeert,Mower2019}.


\begin{figure}[H]
    \centering
    \includegraphics[width=0.8\linewidth]{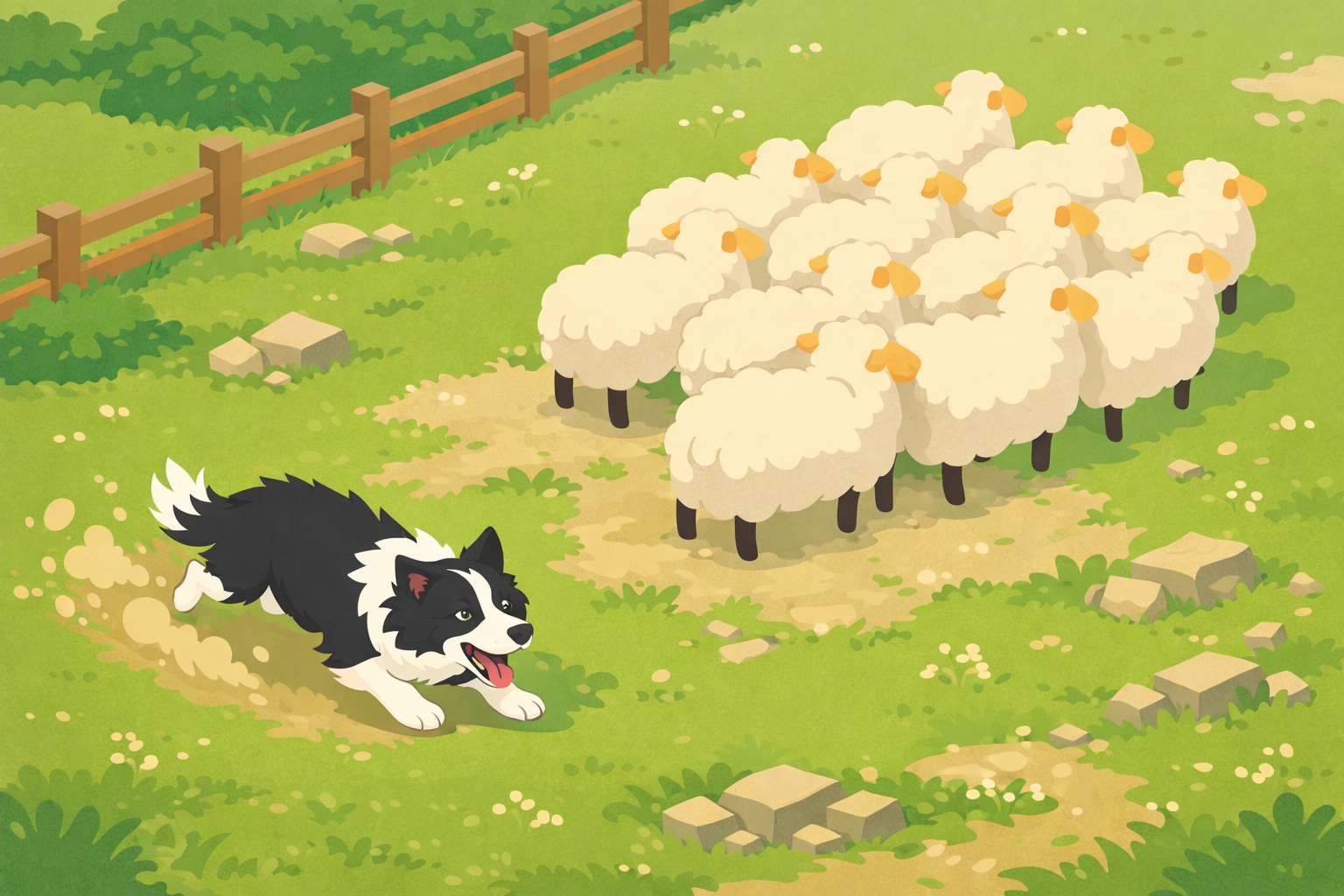}
    \caption{\textcolor{black}{The sheepdog moves on a field in response to commands by the
shepherd. The sheepdog influences the movement of the sheep.}}  
    \label{fig:shepherding}
\end{figure}

We designed a VR shepherding game in which participants control a swarm of agents using a helper agent. The swarm dynamics are instantiated using Reynolds’ Boids local-interaction rules~\cite{Reynolds1987} augmented with a repulsion response to the helper agent. Based on LoRA theory and prior findings that primitive-based/shared-autonomy assistance can improve operator efficiency and workload outcomes relative to continuous low-level teleoperation~\cite{Beer2014,Javdani2018}, we hypothesize:
\textbf{H1}: The shepherding (motion-primitives) condition will reduce mental workload associated with the task, at the cost of reduced direct control granularity.
\textbf{H2}: Task performance will be higher in the shepherding condition, particularly under more challenging scenarios. 

\section{BACKGROUND}
\subsection{Shepherding as Human--Helper-Agent Control}
Shepherding provides a long-standing example of influencing large collectives through a controllable intermediary. Herding dogs enable a single handler to steer mobile flocks that would be difficult to manage via direct human motion alone; this partnership has been shaped through training practices and selective breeding for responsiveness, attention, and ``stock sense''~\cite{Serpell1995,Coppinger2001,MacLean2014,Bray2021}. Border Collies, in particular, are widely cited as highly specialized working dogs selected primarily for herding ability~\cite{Coppinger2001}. Table~\ref{commands} provides a list of commonly used commands in Anglophone countries, as reported by Long et al.~\cite{Long2019}, first formalized by Bennett et al.~\cite{benett} and Coppinger et al.~\cite{Coppinger2001}.

\begin{table}[H]
\caption{Common sheepdog commands and their functional meanings in livestock control.}
\centering
\setlength{\tabcolsep}{6pt}
\footnotesize
\begin{tabularx}{\linewidth}{>{\raggedright\arraybackslash}p{2.3cm} >{\raggedright\arraybackslash}X}
\toprule
\textbf{Command} & \textbf{Functional Meaning} \\
\midrule
Come-bye~\cite{Long2019,Coppinger2001} 
& Flank clockwise around stock to adjust pressure and balance relative to the handler. \\

Away~\cite{Long2019,Coppinger2001} 
& Flank counter-clockwise around stock to adjust pressure and balance relative to the handler. \\

Walk-up~\cite{Coppinger2001} 
& Approach directly toward the livestock to apply forward pressure and initiate movement. \\

Stand/Stop~\cite{Long2019,Coppinger2001} 
& Halt or hold position to remove or modulate pressure on the stock. \\

Look back~\cite{Coppinger2001} 
& Disengage and gather left-behind livestock. \\

That’ll do / Call off~\cite{Coppinger2001} 
& Terminate work and return control focus to the handler. \\
\bottomrule
\label{commands}
\end{tabularx}
\end{table}
Shepherding is hierarchical control: the human issues compact commands, the dog executes positioning behaviors, and the flock responds via local interaction dynamics (e.g., attraction/repulsion)~\cite{Strombom2014}. Modeling and empirical work suggests that effective herding can often be decomposed into ``collect'' and ``drive'' behaviors that yield reliable group-level guidance~\cite{Strombom2014,Long2019}. 
Operationally, the dog influences the flock by modulating pressure, i.e. approaching and positioning relative to the group so that livestock move away from the dog while remaining cohesive with neighbors. By shifting laterally around the flock, the dog changes the group’s balance point enabling turns, containment, and directed motion without contacting or individually commanding animals. Because the dog performs fine-grained adjustments on-the-fly, the human interface remains high-level, using a small command vocabulary suited to dynamic environments~\cite{Coppinger2001,Bray2021}.

In swarm-control terms, pressure corresponds to a localized influence applied at the swarm boundary: the intermediary agent induces a repulsion field that propagates through local interaction rules (e.g., avoidance and cohesion), thereby steering the collective without physically acting on individuals. The balance point can be understood as the goal-relative location on the swarm perimeter where the intermediary’s influence best aligns the swarm’s net motion with the desired direction of travel; moving the intermediary around the perimeter shifts this effective control point and changes the group’s heading~\cite{Strombom2014}. 

\subsection{Directing Robot Swarms: Typical Assumptions}
Robotics has extensively studied the coordination and control of \emph{robot swarms}--engineered collectives whose behavior emerges from distributed controllers and local interactions~\cite{Debie2024SwarmSurvey}. Common approaches include consensus/flocking controllers~\cite{OlfatiSaber2006}, leader--follower or reference-tracking schemes~\cite{Zhu2024SoNS}, and potential-field or artificial-physics formulations~\cite{GaziPassino2011}. In parallel, human-swarm interaction (HSI) research examines how operators can task and steer such swarms via broadcast commands, subgroup selection, parameter tuning, or direct control of designated leaders~\cite{Lyons2025HSIChallenges}.
A key limitation is that most of this literature assumes swarm members are \emph{addressable}: individuals can be programmed, communicated with, or instrumented to execute specified policies. These assumptions fail for ungovernable swarms (e.g., wildlife, fish schools) where individuals cannot be programmed or reliably commanded.







\begin{figure*}
    \centering
    \includegraphics[width=1\linewidth]{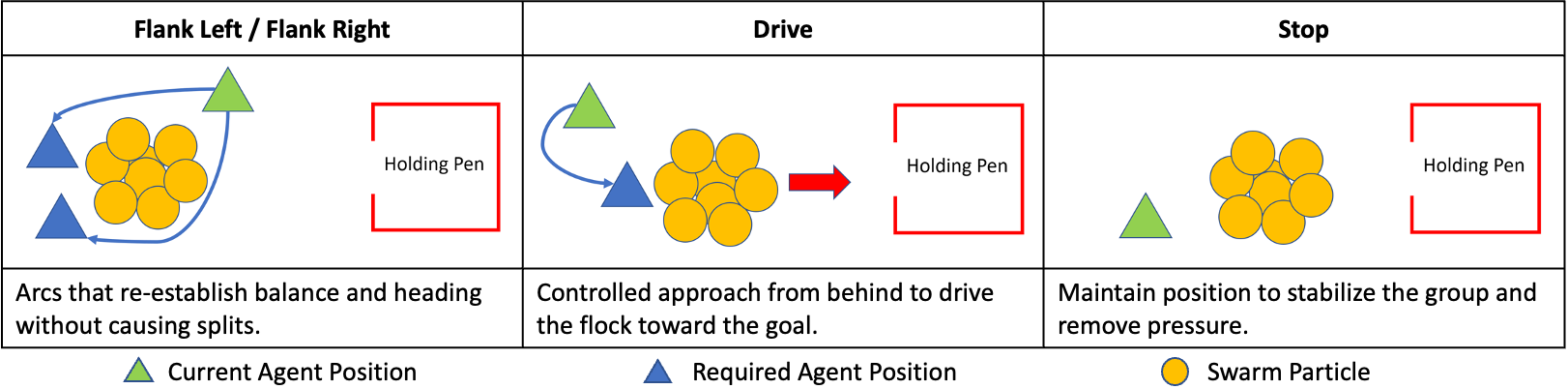}
    \caption{Illustrations to describe shepherding inspired motion primitives implemented in this paper.}
    \label{fig:primitives}
\end{figure*}

\subsection{Bridging the Gap via Helper-Agent Mediation}
This motivates a different control pathway: rather than communicating with swarm members, the human influences collective dynamics through a controllable helper agent that interacts with the swarm in the environment. Shepherding is the canonical example of this mediation strategy.


\section{METHODOLOGY}
To test our hypotheses, a VR shepherding game is designed in which participants use a sheepdog-like helper agent to steer a flock of boids into a goal location. Two human–agent interaction schemes are compared: (i) a teleoperation baseline condition (continuous, low-level control of the helper agent’s position and velocity), and (ii) a shepherding-based motion primitives condition (discrete commands inspired by sheepdog work). Both interfaces are evaluated under two swarm-dynamics scenarios (low vs. high cohesion) within subjects, with interface order, as well as swarm cohesion levels counterbalanced across participants. 

\subsection{Conditions}

Our conditions differ in how the helper agent is controlled. 
\subsubsection{Condition 1: Teleoperation}
For this condition, the participant directly pilots the agent’s planar motion. The left controller’s joystick maps to heading and forward speed of the agent in the ground plane. Neutral stick position corresponds to zero velocity; deflecting the stick away from the participant moves the agent the direction the stick is deflected towards. The agent responds continuously to joystick input, with simple first-order dynamics (bounded speed and turn rate). There is no automatic path planning or assistance; participants must manage approach angles, and path arcs themselves. No explicit “stop” command is required: releasing the joystick allows the agent to effectively come to a “stand” in place, removing pressure from the flock.

\subsubsection{Condition 2: Motion Primitives}
This condition has participants issue discrete button presses that activate predefined shepherding-inspired behaviors (Figure~\ref{fig:primitives}). Once selected, a primitive runs autonomously in closed loop until the participant switches to another command. Thus, participants specify \emph{what} to do (e.g., flank), while the helper agent determines \emph{how} to execute the motion in real time from the evolving flock geometry.
All primitives are computed relative to the flock boundary, approximated by the convex hull at each timestep. The convex hull is the smallest convex polygon enclosing all agents; its construction and derived quantities are defined in Section~\ref{sec:swarm_dynamics}.
\textbf{Flank-L / Flank-R:} The agent moves along a circular arc offset from the convex hull to re-establish balance without cutting through the flock. The system (i) computes the hull and centroid, (ii) defines the flock-pen axis, and (iii) selects a target on the left/right hull-offset arc corresponding to a balance position that steers the flock toward the pen. The target is continuously updated as the hull moves. These come from ``come-bye'' and ``away'' in Table~\ref{commands}.
\textbf{Drive:} This command is derived from the ``walk-up'' command mentioned in Table~\ref{commands}. The agent applies sustained pressure from behind the flock by tracking a target along a rear hull segment opposite the pen. The target slides laterally along this rear arc to address hull bulges and emerging stragglers, distributing pressure across the back perimeter rather than pushing from a single point.
\textbf{Stop:} The agent holds its current position, removing active pressure, until a new primitive is issued.

\subsection{Swarm Dynamics}
\label{sec:swarm_dynamics}
The swarm used in this experiment is based on Reynolds' Boids model~\cite{Reynolds1987}, in which each particle (``boid'') updates its velocity using local interaction rules. Let $\mathbf{x}_i(t)$ and $\mathbf{v}_i(t)$ denote the position and velocity of boid $i$ at time $t$, and let $\mathcal{N}_i$ denote its neighborhood within a fixed perception radius. We compute the standard Boids steering terms: \emph{separation} ($\mathbf{S}_i$), \emph{alignment} ($\mathbf{A}_i$), \emph{cohesion} ($\mathbf{C}_i$), and add an additional term modeling repulsion from the helper agent.

The boid velocity update used in our implementation is:
\begin{equation}
\mathbf{v}_i(t+1) = \mathbf{v}_i(t) + w_s \mathbf{S}_i + w_a \mathbf{A}_i + w_c \mathbf{C}_i + w_r \mathbf{R}_i,
\end{equation}
where $w_s$, $w_a$, $w_c$, and $w_r$ weight separation, alignment, cohesion, and agent repulsion, respectively.

\paragraph{Separation}
\begin{equation}
\mathbf{S}_i = - \sum_{j \in \mathcal{N}_i} 
\frac{\mathbf{x}_j - \mathbf{x}_i}{\|\mathbf{x}_j - \mathbf{x}_i\|}.
\end{equation}

where $\mathcal{N}_i$ denotes the neighborhood of boid $i$ within a fixed perception radius.

\paragraph{Alignment}
\begin{equation}
\mathbf{A}_i = \frac{1}{|\mathcal{N}_i|} 
\sum_{j \in \mathcal{N}_i} \mathbf{v}_j - \mathbf{v}_i.
\end{equation}

\paragraph{Cohesion}
\begin{equation}
\mathbf{C}_i = 
\left( \frac{1}{|\mathcal{N}_i|} 
\sum_{j \in \mathcal{N}_i} \mathbf{x}_j \right) - \mathbf{x}_i.
\end{equation}

\paragraph{Repulsion from the Helper Agent}
To model the helper agent's influence, each boid experiences a distance-dependent repulsive steering term. Let $\mathbf{x}_a(t)$ denote the agent position and define $d_i = \|\mathbf{x}_i - \mathbf{x}_a\|$. Repulsion is applied only within an influence radius $r_a$:
\begin{equation}
\mathbf{R}_i =
\begin{cases}
\left(\frac{1}{d_i} - \frac{1}{r_a}\right)\frac{\mathbf{x}_i - \mathbf{x}_a}{d_i} & \text{if } d_i < r_a,\\
\mathbf{0} & \text{otherwise}.
\end{cases}
\end{equation}
This yields a smooth repulsive field that increases as the boid approaches the agent and vanishes at the influence boundary.

Position updates follow standard kinematic integration:
\begin{equation}
\mathbf{x}_i(t+1) = \mathbf{x}_i(t) + \mathbf{v}_i(t+1) \Delta t.
\end{equation}

Such local interaction rules are sufficient to produce emergent flocking behavior consistent with biological collectives~\cite{Vicsek1995,Couzin2002}. For this experiment, participants interacted with two swarm configurations differing in internal cohesion strength, implemented by varying $w_c$. Higher values of $w_c$ produce tightly aggregated formations resembling schooling fish~\cite{Couzin2002}, whereas lower values yield looser, more diffuse formations resembling avian flocks~\cite{Ballerini2008}.
For the shepherding condition, it is necessary to estimate the spatial extent of the swarm so that the helper agent can generate appropriate flanking arcs. To approximate the swarm boundary, we compute the convex hull of the set of boid positions at each timestep. Computationally, the hull is constructed using a planar convex hull algorithm (in this case, the QuickHull algorithm~\cite{Barber1996}). The algorithm returns an ordered set of boundary vertices $H = \{\mathbf{h}_1, \mathbf{h}_2, \dots, \mathbf{h}_m\},$
forming a convex polygon that encloses all agents.

From this boundary, we compute:
\begin{itemize}
\item The hull centroid:
\begin{equation}
\mathbf{c}_H = \frac{1}{m} \sum_{k=1}^{m} \mathbf{h}_k,
\end{equation}

\item The maximal radial extent:
\begin{equation}
R = \max_{k} \|\mathbf{h}_k - \mathbf{c}_H\|.
\end{equation}
\end{itemize}
These geometric quantities allow the shepherding agent to compute tangential approach trajectories and maintain appropriate stand-off distance relative to the swarm envelope, enabling controlled flanking behaviors similar to biological shepherding strategies~\cite{Strombom2014}.

\subsection{Shepherding Implementation}

The shepherding interface implements three motion
primitives: \emph{Flank Left/Right}, \emph{Drive},
and \emph{Stop}. Each primitive generates a desired
motion for the helper agent based on the current
flock geometry, rather than requiring continuous
user control.

Let $\mathbf{A}(t)$ denote the helper agent's
position, $\mathbf{G}$ the goal location, and
$\mathbf{C}(t)$ the flock centroid. As described
in Section III-B, the flock boundary is approximated
using a convex hull with vertices
$H=\{\mathbf{h}_1,\ldots,\mathbf{h}_m\}$.
The hull centroid and maximum radial extent are
computed as

\begin{equation}
    \mathbf{C}
    =
    \frac{1}{m}
    \sum_{k=1}^{m}\mathbf{h}_k,
    \qquad
    R = \max_k \|\mathbf{h}_k-\mathbf{C}\|.
\end{equation}

The radial extent $R$ provides a conservative
estimate of the flock boundary. Both Flank and
Drive use this quantity to generate target
waypoints outside the convex hull, reducing
the likelihood of the helper entering and
dispersing the flock.

\subsubsection{Flank Left/Right}

The flanking primitive moves the helper agent
around the flock along a circular arc, allowing
it to adjust lateral pressure without passing
directly through the flock. Its inputs are the flock centroid $\mathbf{C}$,
the current agent position $\mathbf{A}$,
the flank clearance $d_f$, and the selected
flanking direction (left or right). First, the radial vector from the flock centroid
to the helper agent is computed:

\begin{equation}
    \mathbf{r} = \mathbf{A}-\mathbf{C}.
\end{equation}

The radial vector is normalized and rotated
by an angular increment $\pm\theta$ around
the centroid, where the sign determines the
flanking direction:

\begin{equation}
    \hat{\mathbf{r}}_{\mathrm{flank}}
    =
    \operatorname{Rot}(\pm\theta)
    \left(
    \frac{\mathbf{r}}{\|\mathbf{r}\|}
    \right).
\end{equation}

Here, $\operatorname{Rot}(\pm\theta)$ represents
a planar rotation about the vertical axis
through the flock centroid.

The desired flank radius $R_f$ is computed
using the convex hull extent $R$ and an
additional clearance $d_f$:

\begin{equation}
    R_f = R+d_f.
\end{equation}

The resulting flank waypoint is

\begin{equation}
    \mathbf{P}_{\mathrm{flank}}
    =
    \mathbf{C}
    +
    (R+d_f)
    \hat{\mathbf{r}}_{\mathrm{flank}}.
\end{equation}

This places the target waypoint outside
the estimated flock boundary. As the agent
and flock move, the radial direction and
waypoint are recomputed, producing a sequence
of targets along the selected flanking arc.
The additional clearance helps prevent
excessive pressure that could disperse
the flock.

\subsubsection{Drive}

The driving primitive positions the helper
agent behind the flock relative to the goal,
allowing the agent's repulsive influence
to guide the flock toward the target region. Its inputs are the flock centroid $\mathbf{C}$,
goal position $\mathbf{G}$, convex hull
extent $R$, and stand-off clearance $d$. First, the unit direction vector from the
flock centroid toward the goal is computed:

\begin{equation}
    \hat{\mathbf{u}}
    =
    \frac{\mathbf{G}-\mathbf{C}}
    {\|\mathbf{G}-\mathbf{C}\|}.
\end{equation}

The desired drive waypoint is positioned
behind the flock along the direction opposite
to the goal:

\begin{equation}
    \mathbf{P}_{\mathrm{drive}}
    =
    \mathbf{C}
    -
    (R+d)\hat{\mathbf{u}}.
\end{equation}

Here, $R$ accounts for the flock's spatial
extent, while $d$ provides additional
clearance beyond the convex hull.
Consequently, the target remains behind
the flock without being placed inside
its estimated boundary. As the helper approaches this position,
nearby boids move away from it according
to the repulsion rule described in
Section III-B, inducing collective motion
toward the goal. The drive waypoint is continuously updated
as the flock centroid and convex hull change,
allowing the helper to maintain its position
behind the moving flock.

\subsubsection{Stop}

The Stop primitive suspends active navigation
and holds the helper agent at its current
position. Unlike Flank and Drive, this command
does not require a new waypoint derived from
the flock geometry. When Stop is issued at time $t_s$, the
holding position is defined as

\begin{equation}
    \mathbf{P}_{\mathrm{stop}}
    =
    \mathbf{A}(t_s).
\end{equation}

The commanded velocity is set to zero:

\begin{equation}
    \mathbf{v}_{\mathrm{cmd}}(t)
    =
    \mathbf{0},
    \qquad t_s \leq t < t_{\mathrm{next}},
\end{equation}

where $t_{\mathrm{next}}$ denotes the time
at which another primitive is activated. This removes active movement of the helper,
allowing the user to observe the flock and
decide on the next command. Although the
agent remains stationary, nearby boids may
continue responding to its repulsive
influence.

\subsubsection{Command Execution}

The motion primitives are activated through
discrete controller inputs. Once selected,
Flank and Drive continuously compute updated
target waypoints using the current flock
geometry. These targets are passed to the
helper agent's navigation controller, which
handles movement toward the specified
positions. The selected primitive remains active until
the participant issues a different command.
Stop suspends the active motion and retains
the agent's current position. Thus, participants specify the intended
behavior while the helper agent autonomously
computes the corresponding motion. The convex-hull-based clearance constrains
the generated waypoints to lie outside the
flock boundary.

\subsection{Task Flow}
\begin{figure*}
    \centering
    \includegraphics[width=1\linewidth]{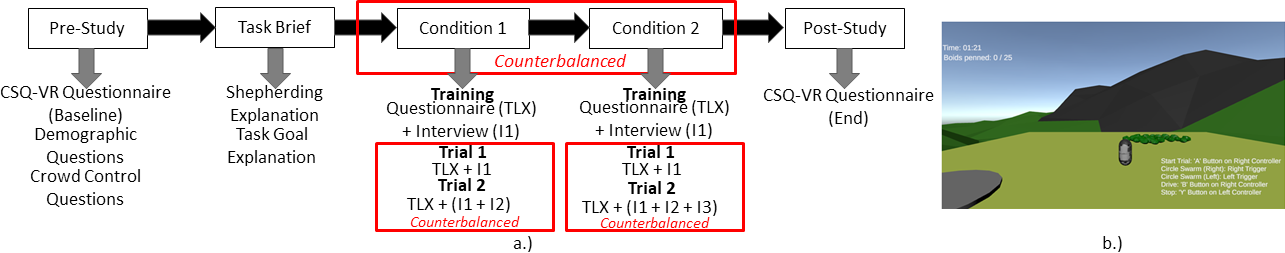}
    \caption{(a) Flow of the experiment. (b) Experiment Setup: Grey capsule represents the agent controlled by user, green particles represent the swarm.}
    \label{fig:taskflow}
\end{figure*}

The VR game was developed using Unity 6.2~\cite{Unity62}, and was deployed on a Meta Quest Pro headset for the experiment. The overall flow of the experiment is described in Figure \ref{fig:taskflow}. Each trial began with 25 boids distributed within a bounded workspace. Participants were instructed to guide a swarm of boids into a designated goal region (the holding pen). Boids entering the pen were removed from the scene, and participants received one point per successfully penned particle. A real-time scoreboard displayed the current number of penned boids, and a countdown timer indicated the remaining time.


\subsubsection{Pre-Trial}
Upon arrival, participants completed an informed consent form followed by a demographic questionnaire. To establish a baseline measure of potential cybersickness symptoms prior to VR exposure, participants completed the Cybersickness Questionnaire for Virtual Reality (CSQ-VR)~\cite{Bouchard2007}. 
Participants were then asked a brief set of interview questions regarding prior experience with swarm management, video games, robotics, or familiarity with shepherding practices. These responses were used to contextualize performance and subjective workload data. 
The exact questions asked during this phase are described below:
\begin{enumerate}
    \item Have you ever had to manage a group of agents before (eg., animals, insects, etc.)?
    \item Have you heard of shepherding before? If so, could you describe what the term means to you please?
\end{enumerate}

\subsubsection{Experimental Trials}
Each condition consisted of three trials, each lasting 90 seconds. Trials were initiated via a standardized trigger screen requiring participant input to begin.
\begin{enumerate}
    \item \textbf{Training Trial:}  
    Participants first completed a training scene designed to familiarize them with the VR environment and the assigned control interface. In this trial, the holding pen was twice the size of that used in subsequent trials to reduce initial difficulty. Participants were encouraged to explore how the flock responded to their control inputs and to refine their strategy before proceeding to scored trials.
    \item \textbf{High Cohesion Trial:}  
    This trial instantiated the high-cohesion swarm configuration. The swarm began dispersed within a bounded workspace, the holding pen was positioned near the participant’s initial viewpoint, and the helper agent began at a fixed starting location. Participants were instructed to herd as many boids as possible into the pen within the time limit.
    \item \textbf{Low Cohesion Trial:}  
    This trial was structurally identical to the high-cohesion trial, with the only difference being the internal swarm cohesion parameter. Spatial layout, pen location, and initial agent positioning were held constant to isolate the effect of swarm dynamics. We expect this trial to be the more difficult trial as the swarm is less likely to stay together.
\end{enumerate}

At the beginning of each trial, all dynamic entities (agent and the boids) are frozen. A start overlay displays a reminder of the control scheme, the current score (initialized to zero), and the trial duration. Participants initiate the trial by pressing the controller trigger, which simultaneously unfreezes all entities and starts the timer.

Once active, participants could use the assigned interaction scheme (teleoperation or motion primitives) to herd the flock toward the pen. Trials concluded automatically when the 90-second time limit elapsed. At trial end, the agent and swarm were frozen in their final positions to preserve the terminal configuration, and the score and timer ceased updating. The system then transitioned to the next trigger screen in the experimental sequence.

Following each trial, participants completed the NASA Task Load Index (NASA-TLX) to assess perceived workload~\cite{Hart1988}. Subsequently, a short semi-structured interview was conducted to capture qualitative impressions of strategy, perceived difficulty, and interaction usability. While a core set of questions (as described below) was asked consistently, the researcher followed relevant lines of inquiry as they emerged:

\begin{enumerate}
    \item In your own words, can you describe how this trial went?
    \item How easy or difficult was it to learn the controls for this condition?
    \item Did you notice any difference between the swarm behavior for all the trials?
\end{enumerate}

\subsubsection{Post-Trial}
After the conclusion of all the trials, participants were engaged in a final interview about their experiences. This was a semi-structured interview, with the following questions:

\begin{enumerate}
    \item For the task you just did, what condition did you prefer, and why?
    \item What would be your ideal control scheme for this task?
\end{enumerate}

Following this interview, they were asked to fill out the CSQ-VR questionnaire again, upon which, they were free to leave the experiment space. The researcher also mentioned that participants were free to ask any questions about the experiment if they so wished.

\subsection{Participant Background}
A total of 24 participants were recruited for the study, primarily from a university population. The sample included individuals from diverse demographic backgrounds. In terms of gender identity, 18 participants self-identified as male, 4 as female, and 2 as non-binary. Participants also reported varied racial/ethnic backgrounds: 12 identified as White/Caucasian, 5 as East Asian, 3 as South Asian, 2 as Black, and 2 as mixed race. Participants fell within two primary age ranges: 20--29 and 30--39. The study protocol was approved by the University's Institutional Review Board (IRB) under an expedited review pathway, and was classified as minimal risk under the University IRB.


\section{RESULTS}
We evaluated task performance, workload, and feedback from the semi-structured interview. Performance and workload were analyzed using $2\times2$ repeated-measures ANOVAs (Interface: Teleoperation vs.\ Shepherding; Cohesion: Low vs.\ High). Answers from the interviews were grouped by themes. For the CSQ-VR survey, no difference was observed in the pre and post-study scores for all of the participants.

\subsection{Task Performance}

\begin{figure}[H]
    \centering
    \includegraphics[width=0.8\linewidth]{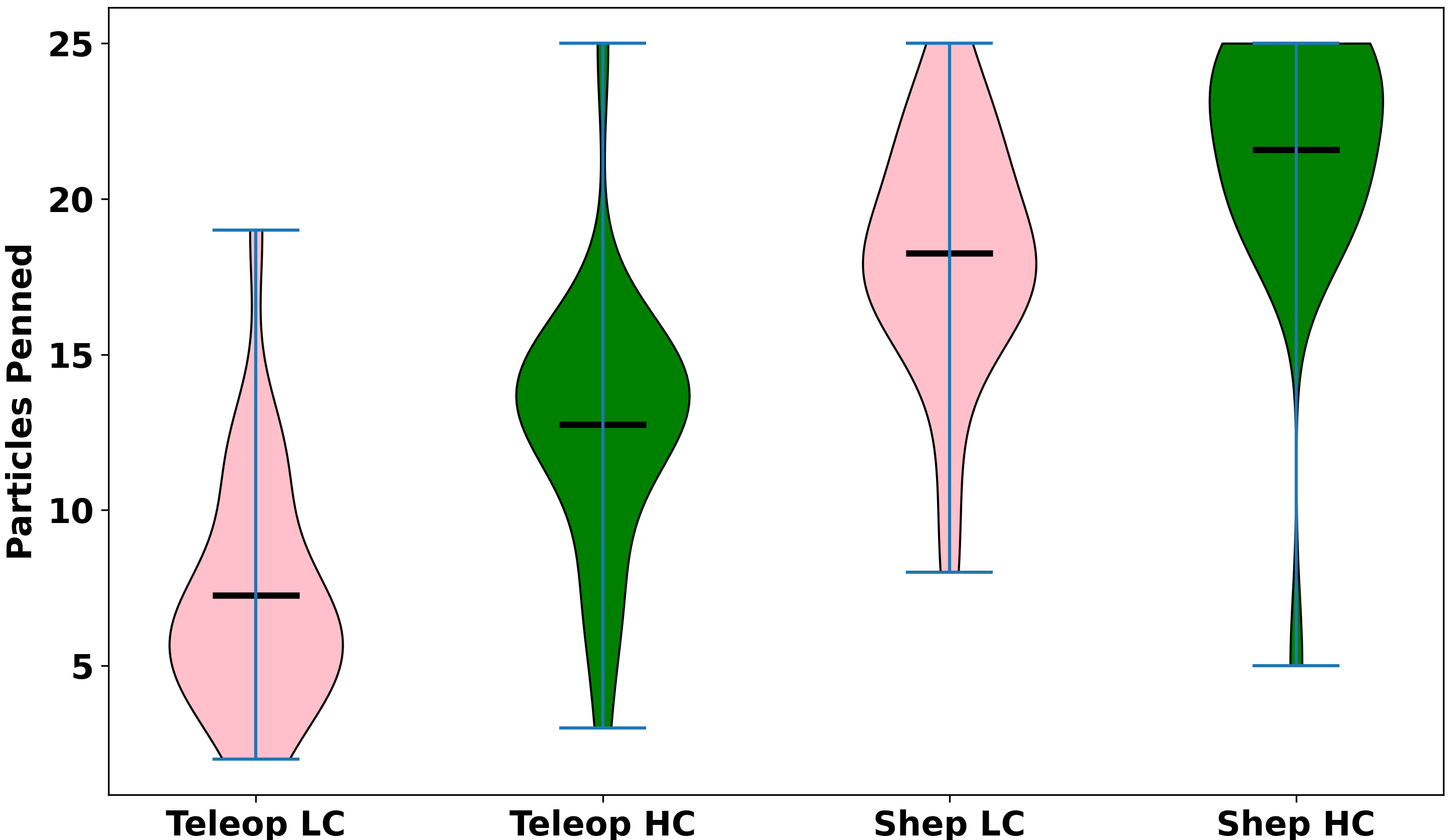}
    \caption{Task Performance Results (LC $\rightarrow$ Low Cohesion; HC $\rightarrow$ High Cohesion; Teleop $\rightarrow$ Teleoperation; Shep $\rightarrow$ Shepherding)}
    \label{fig:performance}
\end{figure}

The primary quantitative performance metric was the number of boids successfully guided into the holding pen by the end of each trial. As described in Section~3.3, each trial began with 25 boids. Participants were instructed to herd as many boids as possible into the pen within the allotted time; each particle that entered the pen contributed one point. The final score therefore served as the operational measure of task performance. Figure~\ref{fig:performance} presents aggregated performance across all conditions.

The analysis showed a main effect of Interface, $F(1,23)=76.55$, $p<.001$. Participants achieved higher scores on average in the shepherding condition (20) than when using teleoperation (10). We also found a main effect of Cohesion, $F(1,23)=85.13$, $p<.001$ with more particles penned in the high cohesion scenario compared to the low cohesion scenario. There was no interaction effect ($p=0.133$). Additionally, we found effect sizes $\eta_{p}^{2}$ of 0.769 and 0.787 for interface and cohesion, respectively using the partial eta squared metric~\cite{Lakens2013}. 
 
 \subsection{Subjective Workload}

Following each trial, participants completed the NASA Task Load Index (NASA-TLX) to assess subjective workload. NASA-TLX is a 21-point scale, that conceptualizes workload as a multidimensional construct spanning \emph{Mental Demand}, \emph{Physical Demand}, \emph{Temporal Demand}, \emph{Performance}, \emph{Effort}, and \emph{Frustration}~\cite{Hart1988,Hart2006}. We computed an overall workload score as the mean of the six sub-scales.



Workload was analyzed using a $2\times2$ Repeated-Measures ANOVA (Interface: Teleoperation vs.\ Shepherding; Cohesion: Low vs.\ High). The analysis showed a significant main effect of Interface, $F(1,23)=33.12$, $p<.001$, effect size $\eta_{p}^{2} = 0.590$. Cohesion and the Interface$\times$Cohesion interaction were not significant. 

\begin{figure} [H]
    \centering
    \includegraphics[width=0.8\linewidth]{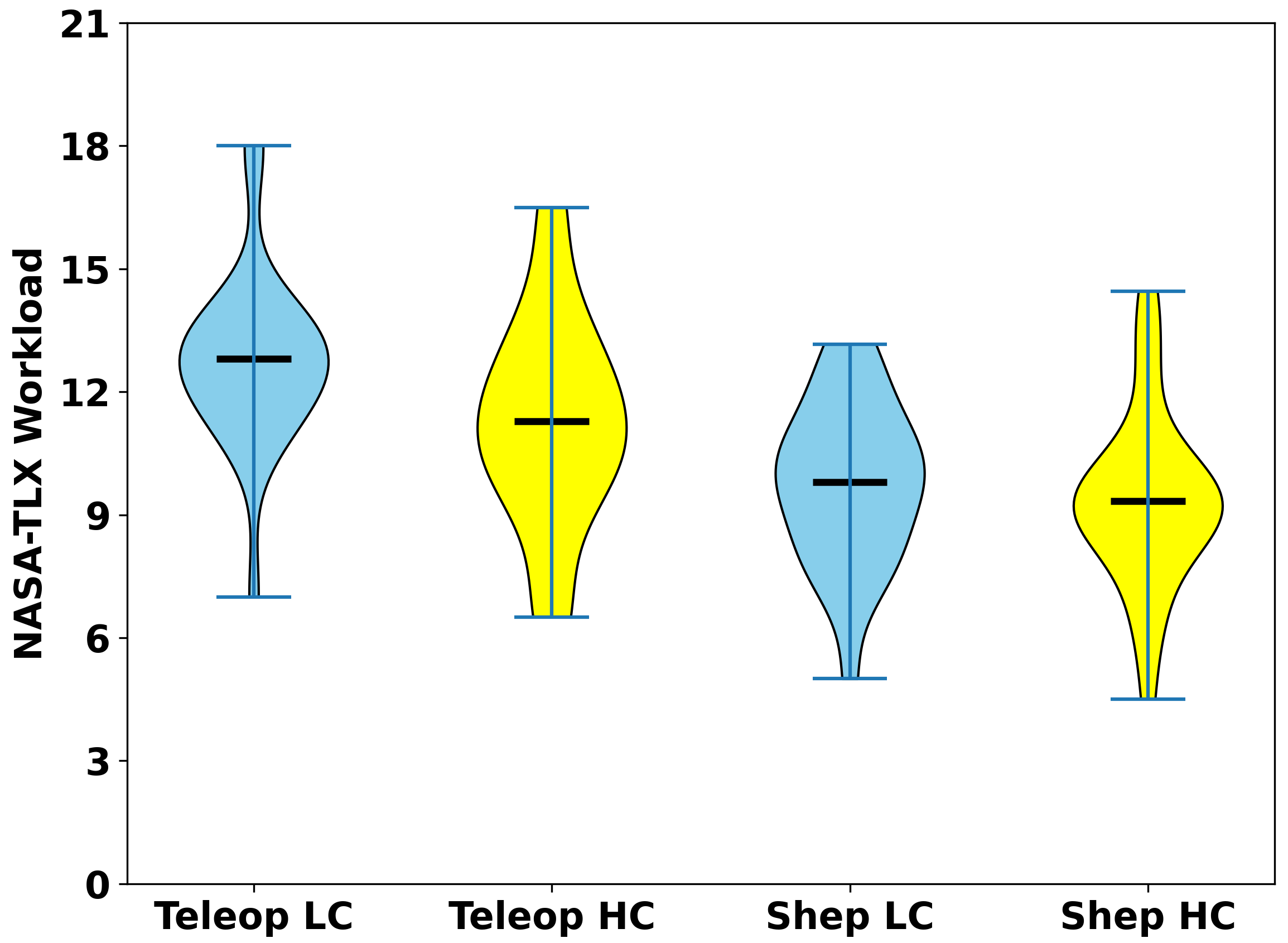}
    \caption{NASA-TLX Results (LC $\rightarrow$ Low Cohesion; HC $\rightarrow$ High Cohesion)}
    \label{fig:workload}
\end{figure}

Figure~\ref{fig:workload} illustrates how the shepherding interface reduces subjective workload (9.66 vs. 11.87). NASA-TLX aggregates multiple components of workload. Autonomy-related benefits may primarily reduce \emph{Mental Demand} and \emph{Effort} while other dimensions (e.g., \emph{Temporal Demand}) remain similar because both conditions impose identical time limits and comparable task pacing~\cite{Hart1988}. In addition, even with motion primitives, participants must maintain situational awareness (e.g., monitoring flock shape and timing command switches) despite reduced low-level control requirements~\cite{Hart2006,Beer2014}. Finally, short, game-like trials can compress subjective ratings, making moderate differences harder to detect even when objective performance diverges~\cite{Hart2006}.

\subsection{\textbf{Open-Ended Reflections}}
Findings from the semi-structured interview questions shed light on how the two conditions were perceived. We extracted questions and themes and provide corresponding snippets from the answers. 

\subsubsection{Preference for Condition}
Of the 24 participants, 14 preferred to use the shepherding commands, with the remaining 10 preferring to use the teleoperation commands. The primary differentiating factor for the two conditions is the amount of control users have over the agent, as well as the mental demand placed on the user during the task. Table~\ref{tab:prefCond} contains snippets from the answers to to this question. 

\begin{table}[h]
\caption{Interview Snippets on Preference for Control Condition}
\label{tab:prefCond}
\centering
\setlength{\tabcolsep}{6pt}
\footnotesize
\begin{tabularx}{\linewidth}{>{\raggedright\arraybackslash}p{0.8cm} >{\raggedright\arraybackslash}X}
\toprule
\textbf{Partici-pant~ID} & \textbf{Snippet, When asked about what condition they prefer, and why: } \\
\midrule
p3 &
`the teleoperation condition, just because I like to have control over the agent at all times' \\

p10 &
`the shepherding condition, because it's easier on the shepherd and harder on the sheepdog' \\

p12 &
`the first (shepherding) condition; it seemed like the dog knew exactly what to do, and when I controlled the dog, I kept messing up' \\


\bottomrule
\end{tabularx}
\end{table}

\subsubsection{Cognitive Load}
Participants reported that the lower-autonomy teleoperation condition afforded finer-grained control over the helper agent, but required sustained attention to trajectory planning, spatial positioning, and continuous motion regulation. In contrast, the shepherding condition with its motion-primitives reduced the need for moment-to-moment control decisions. These subjective accounts align with the quantitative workload findings and are illustrated through interview excerpts presented in Table \ref{tab:cogLoad}.

\begin{table}[h]
\caption{Interview Snippets on Cognitive Load}
\label{tab:cogLoad}
\centering
\setlength{\tabcolsep}{6pt}
\footnotesize
\begin{tabularx}{\linewidth}{>{\raggedright\arraybackslash}p{0.8cm} >{\raggedright\arraybackslash}X}
\toprule
\textbf{Partici-pant~ID} & \textbf{Snippet} \\
\midrule
 p6 &
 \textbf{When asked to describe how the trial went}: `(for the shepherding condition) it went great! It was almost as if the agent knew the best way to move the swarm' \\


p15 &
\textbf{When asked about how demanding the two trials were compared to each other}: `in the first condition (teleoperation), I felt like I was responsible for the dog's movement at all times and I couldn't take a break. the second condition was easier because i was more relaxed' \\

\bottomrule
\end{tabularx}
\end{table}

\subsubsection{Participant Ideas for Control Schemes} 
As part of the post-trial assessment, participants were invited to propose an ideal control scheme after having experienced both interaction conditions. This open-ended prompt was intended to elicit reflections on interface preferences and perceived trade-offs. A majority of participants (14/24) suggested a hybrid approach that combined elements of continuous teleoperation with discrete motion primitives. A smaller subset indicated satisfaction with the shepherding-based motion-primitives condition as implemented. Additionally, a few participants proposed novel control strategies that were not represented in either experimental condition. 
Representative excerpts illustrating these perspectives are presented in Table~\ref{tab:control}.



  

\begin{table}[h]
\caption{Interview Snippets on Control Schemes}
\label{tab:control}
\centering
\setlength{\tabcolsep}{6pt}
\footnotesize
\begin{tabularx}{\linewidth}{>{\raggedright\arraybackslash}p{0.8cm} >{\raggedright\arraybackslash}X}
\toprule
\textbf{Partici-pant~ID} & \textbf{Snippet} \\
\midrule
p5 &
\textbf{When asked what their ideal control scheme for this task would be}: `I like the sheepdog control scheme, but the arcs the dog made around the swarm were too tight sometimes. I would like for there to be an additional button that helps me fine-tune the arc radius' \\


p13 &
\textbf{When asked what changes they would make to the agent to make the task easier}: `I think having multiple helpers instead of a single helper would be good, because I can then send helpers to round up split groups' \\
\bottomrule
\end{tabularx}
\end{table}
\label{sec:RESULTS}

\section{DISCUSSION}
Results obtained from this experiment support our hypotheses or at least parts of them. We discuss these results below and provide actionable implications for HSI.\\
\textbf{Autonomy Allocation and Task Performance} The shepherding condition using motion-primitives yielded higher task performance than the teleoperation condition, supporting the first part of H2. 
We did not find this effect to be stronger in a more challenging condition involving low-cohesion swarm dynamics. 
In contrast, teleoperation performance varied significantly across swarm conditions, suggesting greater sensitivity to environmental instability.
These findings indicate that structured autonomy can improve robustness. By encoding hull-based flanking and distributed rear pressure into motion primitives, the system embedded task-relevant geometric reasoning that participants otherwise had to compute manually. This aligns with shared autonomy literature demonstrating that abstraction at the task level can enhance performance while preserving user agency~\cite{Javdani2018,Gopinath2016}.

\textbf{Cognitive Load and Control Granularity} 
NASA-TLX results indicate reduced workload in the shepherding (motion-primitives) condition, supporting H1. Qualitative data indicate that teleoperation required sustained spatial reasoning, trajectory planning, and continuous regulation of pressure on the swarm. This reflects a control granularity trade-off. Teleoperation provides maximal control authority but increases cognitive demand. Motion primitives reduce granularity by delegating micro-adjustments to the system. Consistent with LoRA theory~\cite{Beer2014}, autonomy improved efficiency by reallocating control from low-level actuation to task-level intent specification. Notably, several participants described the primitive-based agent as a ``helper,'' suggesting that autonomy allocation influences perceived collaboration as well as workload.




\textbf{Hybrid Control} Participants proposed control schemes that allow the user to override motion primitives with direct teleoperation as desired. Such interfaces would preserve the efficiency benefits of abstraction while giving users fine control in cases such as swarm splits or unexpected dispersal.


\textbf{Swarm Fragmentation} We found that our boids-based swarm split into smaller groups in the low cohesion trials. Even if a single boid splits off, it can skew the centroid computation. We modified the base centroid computation algorithm so that sub-groups of less than three members were ignored in the centroid computation. In the field, shepherds often use multiple sheepdogs to prevent their herds from splitting into sub-groups. 


\textbf{Levels of Autonomy}
The two conditions designed for this experiment are chosen to cover the span of the levels of autonomy possible with currently available technology. Additional levels, such as teleoperation with low-level driving assistance (for example, velocity damping or collision avoidance) could have been added to bridge the gap between the two ends of the spectrum used here. We limited our experiment to two conditions to keep it concise and clean.

\textbf{Implications for HSI}
Our results suggest three design implications for HSI:\\
\textit{Encode task structure, not just low-level automation:} The primitives benefited users primarily because they embedded shepherding-relevant structure (e.g., flank geometry, behind-the-flock driving) that users otherwise had to continuously plan and maintain in the teleoperation condition. This was most consequential under low-cohesion dynamics, where instability increased the cost of manual trajectory synthesis. \emph{Implication:} Autonomy should externalize invariants of effective swarm influence (``where to be'' and ``how to apply pressure'') not merely reduce motor effort. 

\textit{Operate on swarm geometry rather than individuals:} By defining control actions relative to the swarm envelope (via convex-hull-based arcs and rear-boundary driving), the interface provided stable control handles despite deformation and splitting. \emph{Implication:} HSI interfaces should expose geometric abstractions (boundary, centroid, heading, rear region) and support boundary-referenced actions that remain meaningful as the swarm changes shape. 

\textit{Support adjustable autonomy with targeted override:} Participants proposed hybrids to handle edge cases (e.g., splits, overly tight arcs). \emph{Implication:} Primitives should be complemented by parameter tuning (e.g., arc radius) and/or temporary direct control, enabling users to intervene according to what works for them. 


\section{LIMITATIONS AND FUTURE WORK}
Our study was conducted in a controlled virtual environment using simulated Boids dynamics. While this type of local interaction models are widely used to study collective behavior~\cite{Reynolds1987,Vicsek1995,Couzin2002}, real-world swarms (e.g., animal groups, or adversarial drone swarms) may exhibit additional behavioral constraints, sensing limitations, and noise. Consequently, future work should validate the proposed motion-primitives framework on embodied robotic platforms.

Additionally, the experiment task involved a single helper agent interacting with a fixed swarm size (25 agents). Real-world scenarios often involve varying swarm sizes and multiple agents. Extending this framework to multi-helper settings and more variable swarms may reveal new coordination dynamics and interface challenges.

The motion-primitives condition implemented a fixed set of discrete behaviors. While this abstraction reduced workload and improved robustness, it may have constrained user expressiveness in edge cases, such as swarm fragmentation. Future research should investigate dynamic autonomy allocation models~\cite{Beer2014}, where users can fluidly transition between primitive-based control and direct teleoperation. Such adaptive autonomy systems may balance efficiency with flexibility, particularly in ungovernable swarms.

Workload was assessed using NASA-TLX~\cite{Hart1988}, a validated subjective instrument. However, subjective workload does not capture fine-grained cognitive processes such as attention allocation or strategy formation. Future studies may incorporate behavioral metrics (e.g., control smoothness) or physiological measures (e.g., eye tracking, heart-rate variability) to more precisely quantify cognitive effort.

Finally, this study focused on short-duration, laboratory-scale tasks. Real-world deployment would require consideration of long-term user adaptation, trust calibration, and failure recovery mechanisms. Investigating how users learn to collaborate with autonomy over time, and how trust evolves under varying swarm volatility, represents an important direction for future HRI research.

\textbf{Swarm Dynamics} The swarm designed for this experiment is based on the Boids principles, and it is assumed that the particles, as well as helpers, are purely kinematic bodies. Factors such as sliding friction, communication latency between controller and agent, etc. are not considered. This is a deviation from technology being used in the field currently, and thus, real-world testing would help validate the findings. Furthermore, parameters for the boids' as well as helper's movement were derived through a mix of testing, as well as taking inspiration from shepherding (by reflecting the movement speeds observed in shepherding). These worked well for the VR task, but real-world deployment may need further fine tuning depending on the swarm's characteristics.

\textbf{Convex Hull Computation} We assume that all swarm particles are perfectly visible for computing the convex hull. In real-life scenarios, occlusions will prevent the real-time location of all swarm members from being known at all times. Additionally, the hull-based primitives introduce an $O(n\log n)$ QuickHull computation per update for $n$ boids. This cost is negligible for our 25-boid VR task, but larger swarms could use lower-rate updates, incremental hulls, and other techniques, instead of recomputing a hull every frame.

\textbf{Workspace} Our experiment workspace is flat and without obstructions. Introducing static obstacles inside the workspace could lead to more realistic trials. Although this idea was not tested as part of this study, it is an avenue worth exploring in future work. Previous work has shown that higher autonomy can reduce operator workload and our findings are consistent with those findings. 

The significance of our study, however, is not simply that autonomy outperforms teleoperation, but that the trade-off between agent autonomy and user's control of the agent
is examined for swarms where individual members cannot be directly controlled. In this context, the key question is how autonomy should be structured to let a human influence the collective indirectly through a helper agent. Our results suggest that shepherding-inspired primitives provide an effective abstraction for manipulating the swarm, improving performance and reducing workload in human-guided control of ungovernable swarms. 



\section*{ACKNOWLEDGMENT}
Generative AI (ChatGPT) was used for organizing the Methodology section for conciseness and during the task design process for help with scripting in Unity. ChatGPT was also used for generating Figure 1.



\bibliographystyle{unsrt}
\bibliography{paper_bibliography.bib} 


\end{document}